\documentclass[lettersize,conference]{IEEEtran}
\usepackage{amsmath,amsfonts}

\usepackage{algorithm}
\usepackage{algpseudocode}
\usepackage{array}
\usepackage[caption=false,font=normalsize,labelfont=sf,textfont=sf]{subfig}
\usepackage{textcomp}
\usepackage{stfloats}
\usepackage{url}
\usepackage{tabularx}
\usepackage{verbatim}
\usepackage{graphicx}
\usepackage{tikz}
\usepackage{multirow}
\usepackage{circledtext}
\usepackage{booktabs}

\usepackage{xcolor}
\usetikzlibrary{arrows.meta, positioning, shapes.geometric}
\usepackage{cite}
\usepackage{authblk}
\newcommand{\figitem}[1]{%
  \raisebox{0.2ex}{(Fig.~\ref{fig:CAD}\textcircled{\scriptsize #1})}}

\begin{document}

\title{Towards Vacuum-Assisted Robotic Needle Interventions: \\From Fluid Aspiration to Tissue Biopsy}

% From Fluid Aspiration to Tissue Biopsy: A Unified Vacuum Assisted Robotic End Effector for Ultrasound Guided Needle Interventions}

\author[1]{Sheena Shabana }
\author[1]{Laxmi Nag Laxminarayan Nagesh}
\author[1]{Marta Vallejo}
\author[2]{Karen Donaldson}
\author[1]{Maria Koskinopoulou}
\affil[1]{Heriot Watt University, Edinburgh, UK}
\affil[2]{University of Edinburgh, Edinburgh,  UK}

\setcounter{Maxaffil}{0}
\renewcommand\Affilfont{\itshape\small}

\maketitle

\begin{abstract}
Robotic ultrasound-guided needle interventions have primarily focused on target localization and accurate needle placement, while the subsequent acquisition of diagnostic material remains comparatively underexplored. Moreover, pleural fluid aspiration and tissue biopsy are typically addressed as separate procedures, despite sharing common requirements for ultrasound guidance, safe access, and sample transport. This work presents a unified robotic framework for ultrasound-guided fluid and tissue extraction using a single vacuum-assisted end effector. The proposed system combines force controlled ultrasound scanning, semantic target identification, automated needle deployment, and procedure specific material acquisition. 
A coaxial needle assembly and shared pump assisted pathway enable direct pleural fluid aspiration, while controlled cannula actuation provides the additional cutting action required for solid-tissue acquisition. End-to-end evaluation using a
Franka Emika Panda robot over 50 phantom trials achieved successful
extraction in 23/25 pleural fluid aspiration trials (92\%) and 22/25
tissue acquisition trials (88\%). Following successful tissue-target
access, acquisition and transport were completed in all 22 trials.

\end{abstract}


\section{Introduction}
Percutaneous needle interventions are widely used for diagnostic
and therapeutic procedures, including biopsy, fluid aspiration,
vascular access, tissue ablation, brachytherapy, and targeted drug
delivery~\cite{cheng2024sensing}. Depending on the procedure, the
needle may deliver a therapeutic agent, drain accumulated fluid, or
retrieve tissue for pathological analysis. Pleural fluid aspiration
and tissue biopsy are two representative material-acquisition
procedures. The former removes excess fluid from the pleural space,
whereas the latter retrieves solid tissue for diagnostic
assessment~\cite{karkhanis2012pleural,bodard2023percutaneous}.
Although these procedures acquire materials with different physical
properties, both require target localisation, identification of a
safe access region, controlled needle deployment, and reliable
sample retrieval.

\begin{figure}[!t]
\centering
\includegraphics[width=\columnwidth]{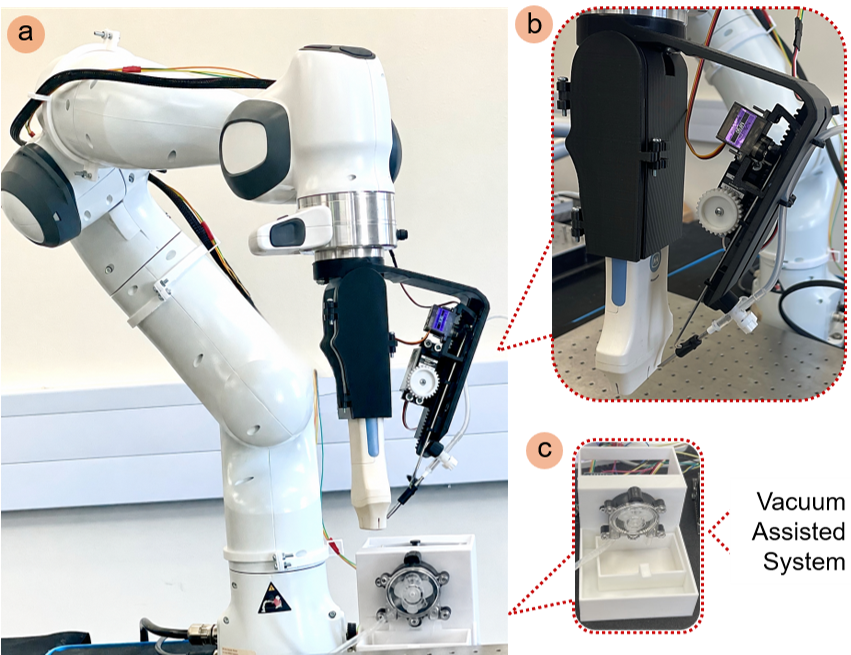}
\caption{Experimental robotic ultrasound-guided extraction platform. 
(a) Franka Emika Panda manipulator equipped with the ultrasound and 
needle end effector. (b) Enlarged view of the unified end effector 
configured for fluid aspiration and tissue biopsy. (c) Peristaltic pump used for 
pump-assisted material transport.}
\label{fig:SystemSetup}
\end{figure}

Needle placement can be guided using computed tomography (CT), magnetic resonance imaging (MRI), or ultrasound (US) imaging
\cite{kulkarni2019review}. Robotic systems operating under CT and
MRI have demonstrated accurate needle alignment and insertion in
thoracic and abdominal interventions, with physiological motion
addressed through breath holding, patient-mounted actuation, and
controlled ventilation
\cite{arnolli2018system, franco2015needle}.
However, CT generally provides intermittent imaging and exposes the
patient to ionising radiation, whereas MRI imposes constraints
related to accessibility, cost, and hardware compatibility.
US instead provides real-time, radiation-free, portable,
and comparatively low-cost imaging~\cite{noble2011ultrasonic},
making it suitable for procedures in which respiration or tissue
deformation may alter the relationship between the target and the
planned needle trajectory~\cite{lawrence2022ultrasound}.

Robotic assistance can improve the stability, precision, and
repeatability of image-guided procedures
\cite{van2012needle}. This has motivated methods
for regulating probe contact and orientation over curved
surfaces~\cite{chen2021automatic}, adapting probe motion to uncertain
surface geometry~\cite{ning2021force}, and identifying scanning
regions affected by acoustic shadowing~\cite{sutedjo2022acoustic}.
For thoracic and abdominal applications, intercostal trajectory
planning~\cite{lei2025toward} and RGB-D-based rib-region
detection~\cite{okuzaki2024rib} have also been investigated to
improve anatomical visibility and access.

Image-based perception has subsequently enabled acquired anatomical
information to be translated into robotic actions. Learning-based
methods have demonstrated multi-organ and lesion segmentation using
task-specific prompts and anatomical priors~\cite{chen2024multi}.
Within robotic needle intervention, segmentation has been used to
localise vascular targets and determine insertion points
\cite{koskinopoulou2023dual}. Other systems have performed needle
insertion following operator-defined target selection
\cite{chen2021ultrasound} or combined reconstructed 3-D images for
planning with real-time 2-D imaging for needle guidance
\cite{lin2023robotic}. Needle localisation has also supported
image-based tracking~\cite{chatelain2013real}, closed-loop steering
and obstacle avoidance~\cite{neubach2009ultrasound}, and improved
visibility under partial occlusion using consecutive frames or
temporal models~\cite{chen2022automatic,goel2024motion}. More
integrated platforms have combined robotic probe manipulation with
needle-tip localisation and trajectory correction
\cite{li2025ultrasound}, while force and tissue-interaction
information has been incorporated to detect breakthrough and reduce
needle overshoot~\cite{liu2025biomechanics}. These developments have
progressively advanced robotic scanning, target localisation, and
controlled needle deployment~\cite{badaan2011does}.

Nevertheless, target access is only one stage of material extraction.
Following insertion, fluid aspiration requires sustained
negative-pressure transport through an open lumen, whereas tissue
biopsy additionally requires mechanical separation of the target
material. Existing robotic systems generally address these functions
separately. Velásquez \textit{et al.} combined pleural-region
detection, needle insertion, and syringe-driven fluid drainage for
autonomous thoracentesis~\cite{velasquez2025ultrasound}. For tissue
sampling, Sajadi \textit{et al.} robotically positioned and triggered
a commercial full-core biopsy instrument
\cite{sajadi2022towards}, while Wang \textit{et al.} integrated
image-based navigation with automated needle deployment and retraction
in thyroid and breast phantoms~\cite{wang2025towards}. Alternative
mechanisms include a magnetically actuated capsule that excises and
retains tissue using a rotating blade and locking mechanism
\cite{cao2025dual}.

Current systems therefore rely on procedure-specific needle mechanisms
and post-access operations, despite sharing upstream requirements for
robotic scanning, anatomical perception, safe access selection, and
needle deployment. The challenge is to preserve these common functions
while adapting only the distal interaction and extraction sequence to
the material being acquired.

In this work, we present a unified robotic system for pleural fluid
aspiration and vacuum-assisted tissue biopsy. As shown in
Fig.~\ref{fig:SystemSetup}, the system integrates force-controlled
robotic scanning, semantic identification of the target and surrounding
bone, target-guided needle deployment, and pump-assisted material
transport. A unified end effector maintains a fixed probe--needle
geometry and incorporates a coaxial needle assembly with independently
controlled insertion depth and cannula configuration. The same assembly
provides an exposed collection lumen for direct fluid aspiration,
whereas, for tissue biopsy, the outer cannula separates the target
tissue before negative pressure transports it through the inner
collection lumen. This acquisition principle is evaluated using
surrogate lesions embedded within an agar-based liver phantom.

The principal contributions of this work are:
\begin{itemize}
    \item The design and realisation of a unified robotic end effector
    that performs both pleural fluid aspiration and tissue biopsy
    using a single coaxial needle assembly. Independent control of
    the assembly insertion depth and relative cannula configuration
    enables procedure-specific penetration, tissue cutting, and
    material collection through a shared pump-assisted extraction
    pathway.

    \item A procedure-aware perception method that
    jointly identifies the acquisition target and surrounding bone,
    constructs a bone-excluded intervention region, and temporally
    verifies its alignment with the mechanically constrained needle
    trajectory before deployment.

    \item An integrated autonomous execution framework coupling
    force-controlled robotic scanning, online target selection,
    motorised needle deployment, procedure-specific material
    acquisition, and safe retraction, validated in 50 phantom trials
    across pleural fluid aspiration and tissue biopsy.
\end{itemize}

% ==========================================
%
% ===========================================

\begin{figure}[!t] 
\centering 
\includegraphics[width=\columnwidth]{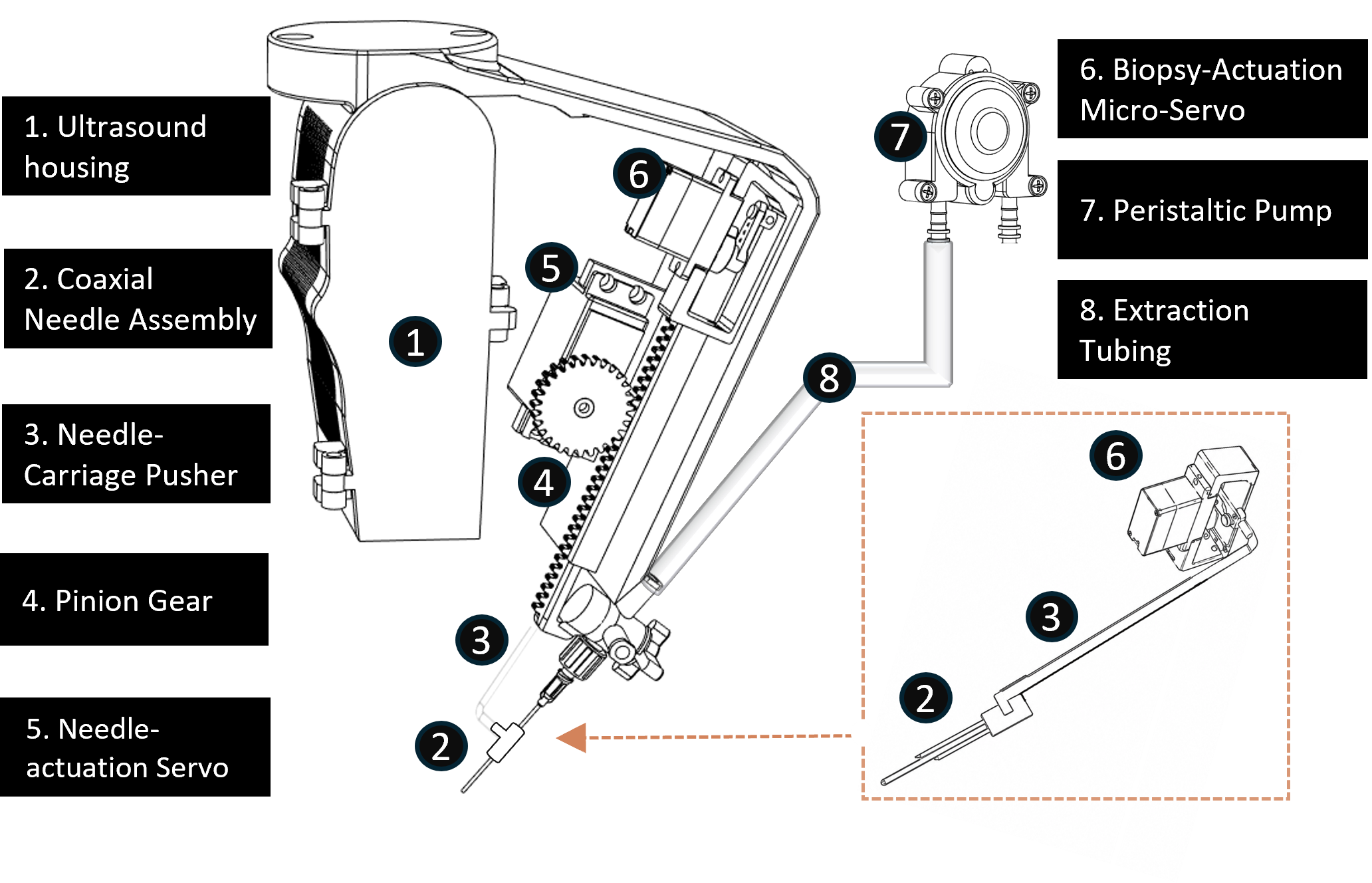} 
\caption{ Mechanical design of the robotic end effector. The
numbered components identify the US housing, coaxial needle
assembly, carriage-based deployment mechanism, independent cannula
actuator, and pump-assisted extraction pathway. The dotted inset highlights
the coupling between the needle assembly, carriage, and
biopsy-actuation micro-servo.} 
\label{fig:CAD} 
\end{figure}

\section{Unified Robotic End Effector} \label{sec:framework}

Pleural fluid aspiration and tissue biopsy share the need for
US-guided access but impose different requirements after
the target is reached. Fluid can be aspirated through an exposed
needle lumen, whereas solid tissue must first be mechanically
separated before it can be transported. 
In this study, the tissue-biopsy configuration
models an intercostal, US-guided liver-biopsy procedure but could be
adapted to other percutaneous soft-tissue biopsy applications with
similar access and material-acquisition requirements.

To accommodate these requirements within a single device, the
mechanical design decouples three functions: (i) maintaining a fixed
probe-needle geometry during US scanning and intervention;
(ii) translating the complete needle assembly along the insertion
axis; and (iii) controlling the relative configuration of the
coaxial cannulas independently of the overall insertion depth.
Fig.~\ref{fig:CAD} shows the resulting end-effector design, while Fig.~\ref{fig:tissue_needle_mechanism}
illustrates the two operating states of the coaxial needle assembly.

% \begin{figure}[!b] 
% \centering 
% \includegraphics[width=\columnwidth]{images/end-effector-design.png} 
% \caption{ Mechanical design of the robotic end effector. The
% numbered components identify the US housing, coaxial needle
% assembly, carriage-based deployment mechanism, independent cannula
% actuator, and pump-assisted extraction pathway. The dotted inset highlights
% the coupling between the needle assembly, carriage, and
% biopsy-actuation micro-servo.} 
% \label{fig:CAD} 
% \end{figure}

\subsection{Design Rationale and Functional Architecture}

The end effector is organised around three mechanically coupled
subsystems. The first is the US guidance structure, formed by
the US housing \figitem{1} and the rigid frame supporting the
needle mechanism. This structure establishes a fixed transformation
between the US imaging region and the needle insertion axis.
The second subsystem provides controlled needle deployment. It
comprises the needle-carriage pusher \figitem{3}, pinion gear
\figitem{4}, and needle-actuation servo \figitem{5}. Together, these
components translate the complete coaxial needle assembly
\figitem{2} towards or away from the intervention target.
The third subsystem controls sample acquisition and transport. The
biopsy-actuation micro-servo \figitem{6} changes the relative axial
position of the two cannulas, while the peristaltic pump
\figitem{7} and extraction tubing \figitem{8} establish a continuous
material pathway between the distal needle opening and the collection
container.
This functional separation allows the overall needle position and
the distal needle configuration to be controlled independently.
Consequently, the insertion depth can be maintained while the needle
transitions between its penetration and collection states. The same
mechanical architecture can therefore support both fluid aspiration
and tissue biopsy without exchanging the needle assembly or modifying
the US guidance structure.

\subsection{Needle Deployment and Cannula Actuation}

The complete coaxial needle assembly \figitem{2} is mounted on the
needle-carriage pusher \figitem{3}, which translates along a linear
guide aligned with the prescribed insertion axis. The
needle-actuation servo \figitem{5} drives the pinion gear
\figitem{4}, which engages with the linear rack integrated with the
needle carriage. Rotation of the pinion is therefore converted into
linear displacement of the carriage and the attached needle assembly.

This rack-and-pinion transmission provides the primary deployment
motion of the end effector. Forward servo rotation advances the
complete coaxial needle assembly towards the selected intervention
region, whereas reverse rotation retracts it following sample
collection. Because both cannulas are supported by the translating
carriage, this motion changes the overall insertion depth without
altering their relative axial configuration.

A second actuation stage is mounted on the needle carriage to control
the distal needle configuration. The biopsy-actuation micro-servo
\figitem{6} is mechanically coupled to the outer-cannula slider
through the connecting linkage shown in Fig.~\ref{fig:CAD}. Actuation of this
servo translates the 10G outer cutting cannula relative to the 12G
inner collection cannula. This differential motion is independent of
the primary rack-and-pinion translation and enables the needle to
switch between the two states shown in Fig.~\ref{fig:tissue_needle_mechanism}.

The mechanical architecture thus provides two independently actuated
linear motions. The primary motion determines the insertion depth of
the complete coaxial needle assembly, while the differential motion
determines whether the sharpened outer cannula or the cylindrical
inner collection cannula forms the distal leading element.

\subsection{Coaxial Needle and Extraction Principle}

The coaxial needle assembly comprises a sharpened 10G outer cutting
cannula and a cylindrical-tip 12G inner collection cannula. The
smaller 12G cannula is positioned within the lumen of the 10G outer
cannula, allowing relative axial displacement while maintaining a
common insertion axis. The outer cannula provides the sharpened edge
required for penetration and tissue cutting, whereas the inner
cannula provides the distal collection opening and the lumen used for
material extraction.

\begin{figure}[t]
    \centering
    \includegraphics[width=\columnwidth]{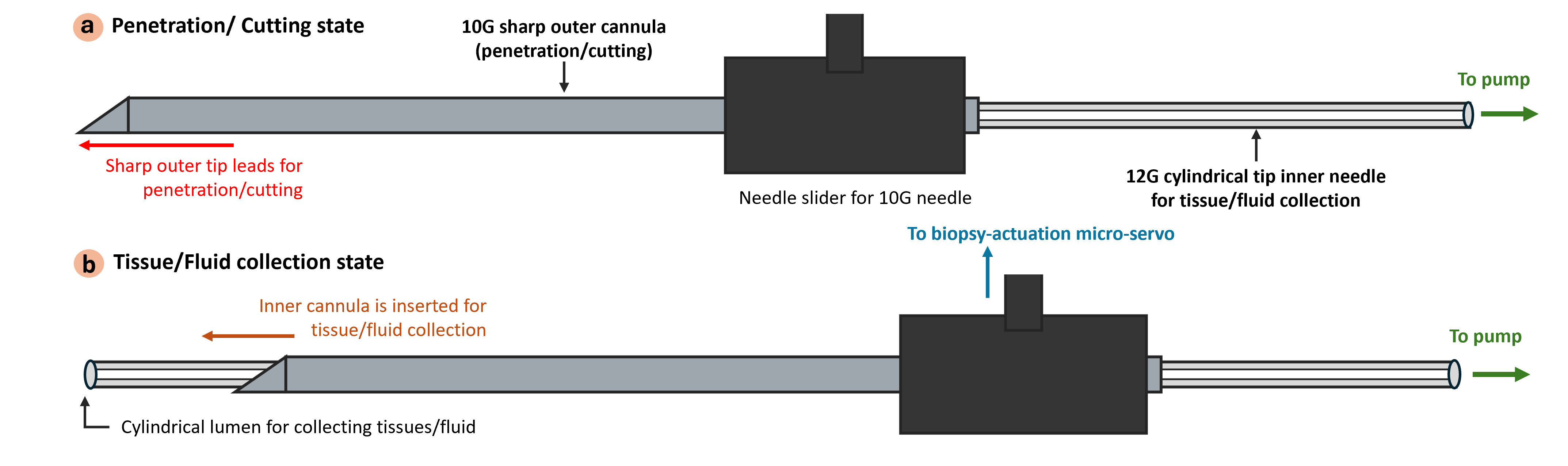}
    \caption{Operating principle of the coaxial needle assembly:
(a) penetration state, in which the sharpened 10G outer cutting
cannula forms the leading element; and (b) collection state: 
the outer cutting cannula is retracted to expose the cylindrical
distal opening and lumen of the 12G inner cannula.}
    \label{fig:tissue_needle_mechanism}
\end{figure}

In the penetration/cutting state shown in Fig.~\ref{fig:tissue_needle_mechanism}(a), the 10G outer
cutting cannula extends beyond the distal end of the 12G inner
collection cannula. Its sharpened tip therefore forms the leading
element of the assembly and provides the cutting action required
during initial access or tissue penetration.

In the collection state shown in Fig.~\ref{fig:tissue_needle_mechanism}(b), the outer cutting cannula
is retracted proximally by the biopsy-actuation micro-servo
\figitem{6}. This exposes the cylindrical distal opening of the inner
collection cannula while the overall insertion depth is maintained.
The inner cannula is connected proximally to the peristaltic pump
\figitem{7} through the extraction tubing \figitem{8}, producing a
continuous extraction pathway from the target region to the
collection container.

For pleural-fluid aspiration, the needle is maintained in the
collection state after access to the effusion region, allowing fluid
to be drawn directly through the inner collection lumen. For tissue
biopsy, the outer cutting cannula is first actuated to penetrate and
separate the target tissue and is then retracted to expose the inner
collection opening. Pump-generated negative pressure subsequently
draws the separated tissue material into the collection lumen and
through the downstream extraction pathway.
Accordingly, the distinction between the two procedures is achieved
through the sequence of the same two mechanical actions: relative
cannula actuation and pump-assisted extraction.

% \begin{figure}[t]
%     \centering
%     \includegraphics[width=\columnwidth]{images/NEEDLE_DESIGN.png}
%     \caption{Operating principle of the coaxial needle assembly:
% (a) penetration state, in which the sharpened 10G outer cutting
% cannula forms the leading element; and (b) collection state: 
% the outer cutting cannula is retracted to expose the cylindrical
% distal opening and lumen of the 12G inner cannula.}
%     \label{fig:tissue_needle_mechanism}
% \end{figure}

% ============================================================

\section{US Guided Intervention}
\label{sec:perception}

The perception module identifies the intervention target during
robotic scanning and determines whether it provides a valid region
for needle access. Before execution, the system is configured for
either pleural fluid aspiration or tissue biopsy. This procedural
input determines the segmentation model and target class used during
inference: pleural fluid region is selected for fluid aspiration, whereas
the embedded tissue structure is selected for biopsy. Bone is treated
as an anatomical exclusion constraint in both procedures.

B-mode images were acquired from the phantom using the Sonon 300L linear US
scanner retained within the end-effector housing
\figitem{1}. A predefined region of interest
corresponding to the B-mode imaging window was extracted from each
frame, and resized to $256\times256$ pixels. {\color{black}Two procedure-specific US datasets were used to develop the
perception models. The pleural aspiration dataset comprises 194
US images containing pleural fluid and rib structures, with
pixel-wise annotations of background, pleural fluid region, and bone. The
tissue acquisition dataset comprises 66 US images acquired from
the tissue phantom and annotated for background, tissue, and bone. All images were manually annotated at the pixel level using
LabelMe \cite{russell2008labelme}. The pleural aspiration dataset was partitioned into 154 training, 40 validation, and 40 held-out test images. The tissue acquisition dataset was partitioned into 146 training, 40 validation, and 40 held-out test images.

Training images were augmented using horizontal flipping and rotations
within $\pm10^{\circ}$.  Three encoder-decoder segmentation architectures, U-Net, Attention U-Net, and U-Net++, were evaluated on both datasets to determine their suitability for the two
perception tasks. For pleural aspiration, each architecture was trained to segment background, pleural fluid region, and bone, whereas for tissue acquisition, the corresponding classes were background, tissue, and bone. This procedure specific formulation enables the relevant intervention target to be segmented while retaining bone as a common anatomical constraint.}
% Three different U-Net models, U-Net, Attention U-Net, U-Net++ were used to train dataset to analyze which model can perform on the dataset better. Attention U-Net was used to segment background, pleural effusion, and bone for fluid aspiration, whereas U-Net++ was used to segment background, tissue, and bone for tissue acquisition.The procedure-specific models account for differences in target appearance and spatial distribution while retaining bone segmentation as a common anatomical constraint. 
The three models were trained with a learning rate of $1\times10^{-4}$, a batch size of 8, and a maximum of 100 epochs. A combined cross-entropy and Dice loss was used to address class imbalance. For each architecture, the model checkpoint with the highest mean foreground Dice score on the validation set was retained.

% During online execution, the procedure-specific model generates
% pixel-wise target and bone masks from each acquired B-mode frame.
% Connected-component filtering and morphological processing are then
% applied before procedure-specific intervention-region validation. In
% both configurations, the predicted bone mask is used as an anatomical
% exclusion constraint such that target regions overlapping the
% bone-exclusion region are rejected.

% For pleural fluid aspiration, the candidate intervention mask is
% obtained from:

% \begin{equation}
% M_{\mathrm{asp}}
% =
% M_{\mathrm{effusion}}
% \setminus
% \mathcal{D}_{\mathrm{asp}}(M_{\mathrm{bone}}),
% \label{eq:asp_region}
% \end{equation}
% % 
% where $M_{\mathrm{effusion}}$ and $M_{\mathrm{bone}}$ denote the
% predicted pleural-effusion and bone masks, respectively, and
% $\mathcal{D}_{\mathrm{asp}}(\cdot)$ denotes the morphological dilation
% used by the aspiration detection algorithm. The resulting mask is
% evaluated using the aspiration-specific geometric acceptance criteria.
% To suppress deployment due to transient predictions, a valid
% intervention region must be detected for $N_f=5$ consecutive frames.
% Once this temporal criterion is satisfied, the detected frame is
% retained, the intervention status is updated, and the needle deployment
% and aspiration sequence is initiated.

\begin{figure*}[t!] 
    \centering
    \includegraphics[width=\textwidth]{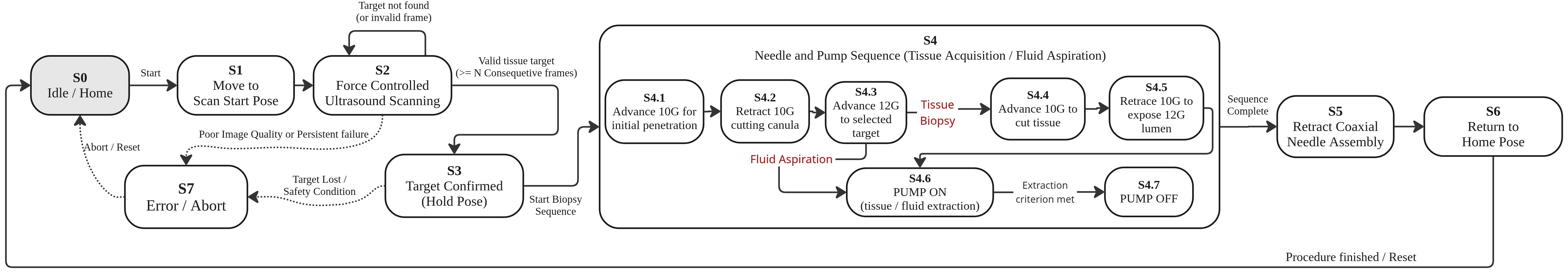}
    \caption{State machine for autonomous pleural fluid aspiration and
tissue biopsy.}
    \label{fig:state_diagram}
\end{figure*}

For pleural fluid aspiration, effusion and bone components smaller than
400 and 120 pixels, respectively, are removed. The bone mask is dilated
using a $15\times15$-pixel structuring element, and the candidate
intervention region is defined as
\begin{equation}
M_{\mathrm{asp}}
=
M_{\mathrm{effusion}}
\setminus
\mathcal{D}(M_{\mathrm{bone}}),
\label{eq:asp_region}
\end{equation}
with components smaller than 400 pixels subsequently removed. A region
is accepted when the effusion and candidate-region areas exceed 15\%
and 20\% of the B-mode ROI, respectively, and its largest component has
a minimum width and height of 150 and 80 pixels. These conditions must persist for $N_f=5$ consecutive frames
before needle deployment is initiated.

For tissue acquisition, tissue and bone components smaller than 300 and
100 pixels, respectively, are removed. The bone mask is dilated using a
$15\times15$-pixel structuring element, and the bone-constrained tissue
region is defined as
\begin{equation}
M_{\mathrm{tissue}}^{\mathrm{BC}}
=
M_{\mathrm{tissue}}
\setminus
\mathcal{D}(M_{\mathrm{bone}}),
\label{eq:tissue_region}
\end{equation}
with components smaller than 300 pixels subsequently removed. A target
is accepted when the tissue and bone-constrained tissue regions occupy
at least 5\% and 3\% of the B-mode ROI, respectively, and the largest
valid component has a minimum width and height of 80 and 40 pixels. A 5 seconds trigger cooldown is imposed after activation to prevent rapid repeated triggering.

\section{System Integration and Autonomous Execution}
\label{sec:system_integration}

The complete experimental platform is shown in
Fig.~\ref{fig:SystemSetup}. The developed end effector was mounted
on a Franka Emika Panda 7-DoF manipulator and integrated with the
US perception and procedure control modules through ROS.
The Sonon 300L scanner provides the B-mode image stream to an NVIDIA
Jetson Nano, where image pre-processing, segmentation, and
intervention region validation are performed. The robot control
workstation executes the scanning trajectory, commands the two
end effector actuators, controls the peristaltic pump, and coordinates
the procedure according to the state machine in
Fig.~\ref{fig:state_diagram}.

Robotic scanning was performed using the Franka Cartesian impedance
controller to maintain compliant contact between the US probe
and the phantom. A PI force-control loop modified the commanded
end-effector position along the local surface-normal direction using
the difference between the desired and measured contact forces. This
normal correction was superimposed on the scanning trajectory,
allowing the robot to follow the phantom surface while regulating the
probe contact force at $0.70$\,N. 
The Cartesian translational and rotational stiffness were set to
$K_t=\mathrm{diag}(1500,1500,300)$\,N/m and
$K_r=\mathrm{diag}(30,30,30)$\,Nm/rad, respectively, while
the PI gains were set to
$K_p=5\times10^{-4}$\,m/N and
$K_i=1\times10^{-4}$\,m/(N\,s).

The accessible surface was represented by a parabolic profile derived
from the measured phantom geometry. A scanning area of
$0.125\times0.105$\,m was covered using four parallel passes executed
at $0.0025$\,m/s. For each commanded position along a pass, the
surface model provided the corresponding height and local gradient.
The probe height was updated to follow the modeled surface, while its
orientation was adjusted to maintain the US transducer
approximately normal to the local surface. The force controller
compensated for residual differences between the modeled and actual
phantom geometry.

\begin{figure}[!t]
\centering
\includegraphics[width=\columnwidth]{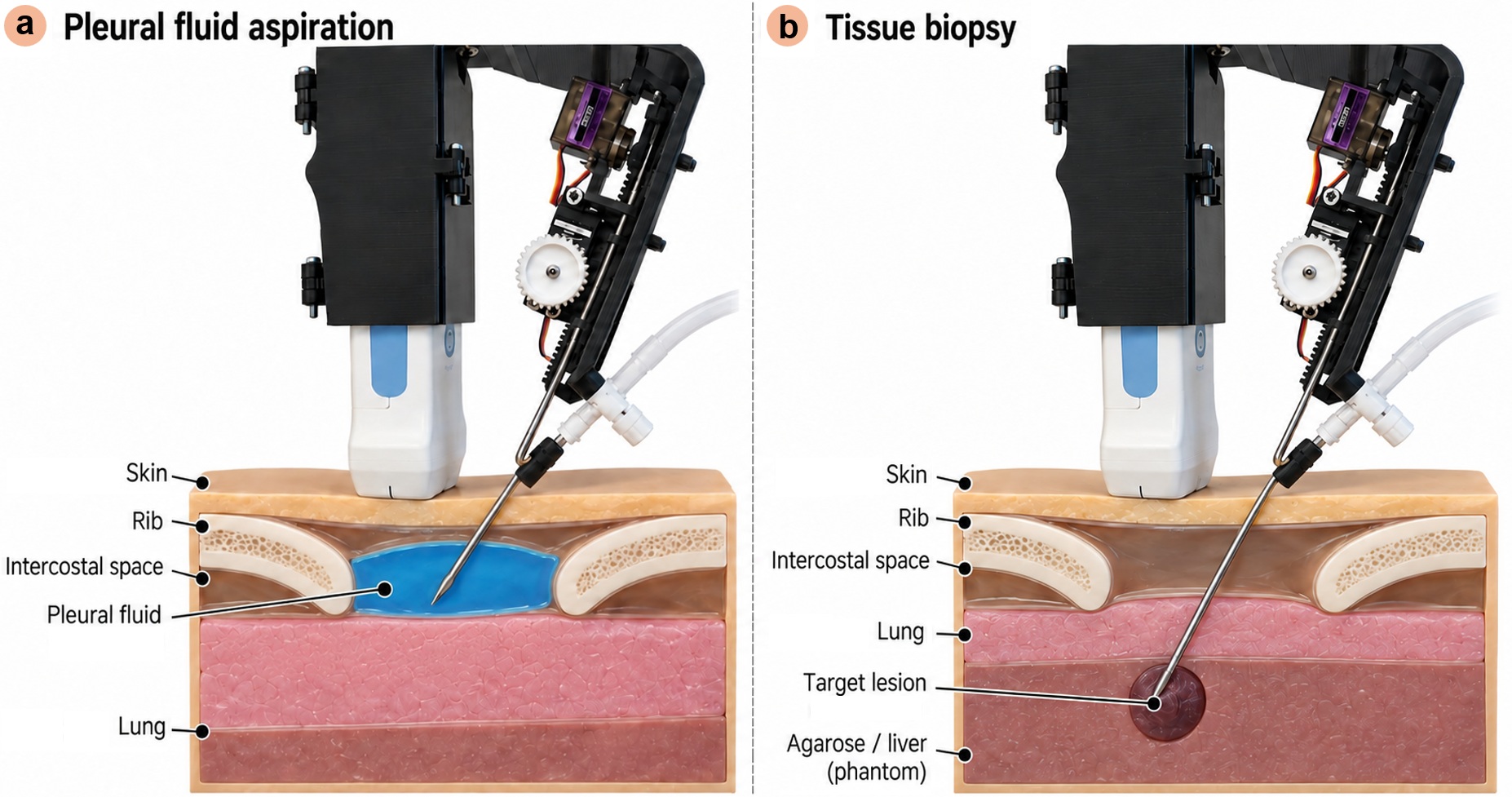}
\caption{Schematic of the two intervention configurations evaluated using a thoracentesis phantom containing an agar-based liver model and end-effector geometry: (a) pleural fluid aspiration, in which the coaxial needle is deployed through the intercostal space into the pleural fluid region; and (b) tissue biopsy, in which the needle follows the same intercostal access geometry to reach a surrogate lesion embedded within the agar-based liver phantom.}
\label{fig:integrated_system}
\end{figure}

A Kyoto Kagaku MW4 thoracentesis phantom was used to
reproduce the principal anatomical constraints associated with
intercostal access. The experiments used the mid-axillary puncture
unit, which represents the sixth to ninth ribs and incorporates
skin-mimicking and intercostal soft-tissue layers, a pleural
compartment, lung, and diaphragm structures. The puncture unit
measured approximately $160\times70\times210$\,mm. Along the
selected insertion region, the combined thickness of the skin and
superficial soft tissue was approximately
$15$\,mm, while the clear spacing between adjacent rib
boundaries was approximately $20$\,mm. These dimensions
defined the available access window and the minimum needle travel
required to reach the internal target.

For pleural fluid aspiration (Fig.~\ref{fig:integrated_system}(a)), the pleural compartment was filled
with $350$\,mL of water, producing an US visible
fluid region approximately $30$\,mm beneath the scanning
surface. The prescribed extraction volume was
$280$\,mL per trial. Pump actuation was terminated when
this volume was collected or when the aspiration duration reached
$155$\,seconds.

\begin{figure}[t!]
    \centering
    \includegraphics[width=\columnwidth]{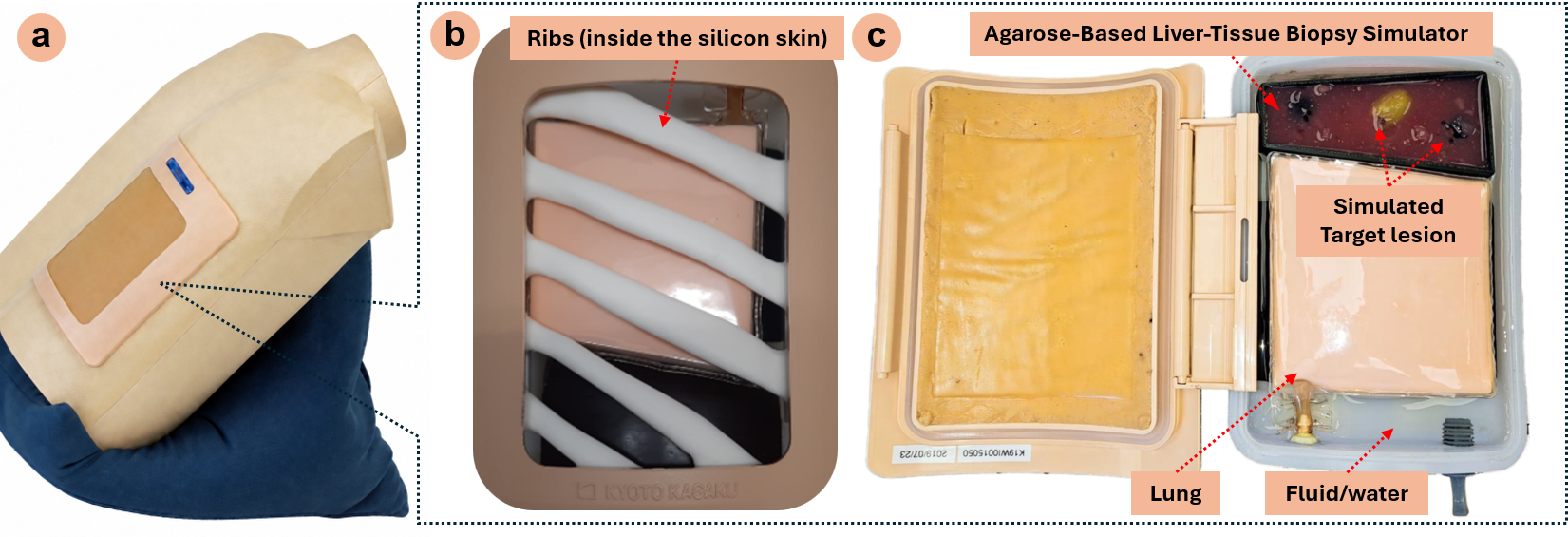}
    \caption{Phantom model used for experimental evaluation:
(a) Kyoto Kagaku MW4 thoracentesis phantom with the mid-axillary
puncture unit; (b) rib structures beneath the skin-mimicking access
surface; and (c) modified internal configuration containing the
custom agar-based liver model with embedded target lesions,
together with the retained lung and fluid compartments.}
    \label{fig:phantom_design}
\end{figure}

For tissue biopsy (Fig.~\ref{fig:integrated_system}(b)), the diaphragm insert was replaced by the custom
agar-based liver model shown in Fig.~\ref{fig:phantom_design}. The model was prepared using an agar-based mixture comprising 100 mL of water and approximately 0.5 - 0.75 teaspoons of agar powder and measured approximately $140\times60\times35$\,mm. Two grapes and four olives were embedded as US visible solid targets. The grapes had a maximum dimension of approximately $20\pm3$\,mm, while the olives had a mean maximum dimension of approximately $15\pm2$\,mm. The targets were positioned approximately $15$--$30$\,mm below the model surface. The custom model was placed within the puncture unit while retaining the original skin-mimicking layer and rib-containing cover. This configuration enabled pleural fluid aspiration and tissue biopsy to be evaluated through the same external scanning surface and intercostal access geometry.

\subsection{Autonomous Execution}

The autonomous procedure begins in the idle/home state (S0) shown in
Fig.~\ref{fig:state_diagram}. Following a start command, the robot
moves to the predefined scan-start pose (S1) and activates
force-controlled scanning (S2). US frames are processed
online using the procedure-specific perception method described in
Section~\ref{sec:perception}. Scanning continues until a
bone-free target region aligned with the fixed needle trajectory is
detected for $N_f$ consecutive frames.

Once the target-validity condition is satisfied, the tangential
scanning motion is stopped and the system enters the
target-confirmed state (S3). The Franka controller maintains the
selected pose and compliant probe contact while the end-effector
controller executes the needle-and-pump sequence (S4). Needle motion
within this state follows the coaxial mechanism described in
Section~\ref{sec:framework} and illustrated in
Fig.~\ref{fig:tissue_needle_mechanism}.

For pleural fluid aspiration, the collection cannula is positioned
within the confirmed effusion region, and the pump is activated
without a tissue-cutting action. For tissue biopsy, the outer cannula
is additionally advanced and retracted to separate the target tissue
and expose the inner collection lumen before pump activation. The pump is subsequently activated for $30~s$ to transport the acquired tissue through the 12G lumen to the collection chamber. The two procedures, therefore, share target-guided needle deployment and pump-assisted transport but differ in the distal cannula actuation performed before extraction.

When the procedure-specific extraction criterion is reached, the
pump is stopped, and the coaxial needle assembly is retracted (S5).
The manipulator then returns to its home pose (S6). Failure to obtain
a valid target, loss of the target before deployment, or violation of
a predefined safety condition transfers the system to the error/abort
state (S7), preventing further needle or pump actuation.

% \newpage
% -----------------------------------------------------------------------

\section{Experimental Evaluation}

{\textcolor{black}{The experimental platform described in
Section~\ref{sec:system_integration} was used to evaluate the proposed system under two intervention configurations: pleural fluid aspiration and tissue acquisition. The evaluation was performed at two levels.
First, the US perception module was evaluated offline to characterize target and bone segmentation accuracy and spatial
localization performance. Second, the complete robotic pipeline was evaluated to quantify target-guided needle deployment and material
extraction performance.}}

\subsection{US Segmentation Evaluation}
The three segmentation architectures were evaluated on the held-out test sets for the two intervention configurations. 
All segmentation metrics and class-wise DSC values are reported in the tables as the mean and standard deviation across images in the corresponding test set. Performance was
assessed separately for the procedure-specific target---pleural
effusion for aspiration and the embedded tissue structure for
biopsy---and for bone, which represents the anatomical exclusion
constraint. Table~\ref{tab:segmentation} reports the Dice score for
each class, their mean Dice similarity coefficient (DSC), and the 95th percentile Hausdorff distance (HD95).
Higher DSC scores indicate greater agreement with the ground-truth
regions, whereas a lower HD95 indicates more accurate target-boundary
localisation.

For pleural fluid aspiration, Attention U-Net provided the most
balanced performance across the target and bone classes. Although U-Net++
achieved a marginally higher pleural-effusion DSC, its lower bone DSC
reduced its suitability for bone-aware intervention planning.
Attention U-Net was therefore selected, achieving the highest mean DSC
(0.817) and lowest target HD95 (35.22 pixels).

For tissue acquisition, U-Net++ achieved the highest tissue DSC
(0.883) and mDSC, together with the lowest target HD95. Although
U-Net produced a slightly higher bone DSC, U-Net++ provided more
accurate target delineation and was therefore selected for the
end-to-end tissue experiments.

Spatial localisation of the tissue target was additionally evaluated
using the centroid error of the largest connected tissue component:

\begin{equation}
E_{\mathrm{loc}} =
\sqrt{(x_{\mathrm{pred}}-x_{\mathrm{GT}})^2 +
      (y_{\mathrm{pred}}-y_{\mathrm{GT}})^2},
\end{equation}
where $(x_{\mathrm{pred}},y_{\mathrm{pred}})$ and
$(x_{\mathrm{GT}},y_{\mathrm{GT}})$ denote the predicted and
ground-truth centroids, respectively. U-Net++ achieved the lowest
centroid error of $16.91\pm25.38$ pixels, further supporting its
selection for tissue acquisition. The relatively large variation
indicates that localisation accuracy was influenced by differences
in target visibility, shape, and acoustic shadowing across the test
images.

\begin{figure}[!h]
\centering
\includegraphics[width=\columnwidth]{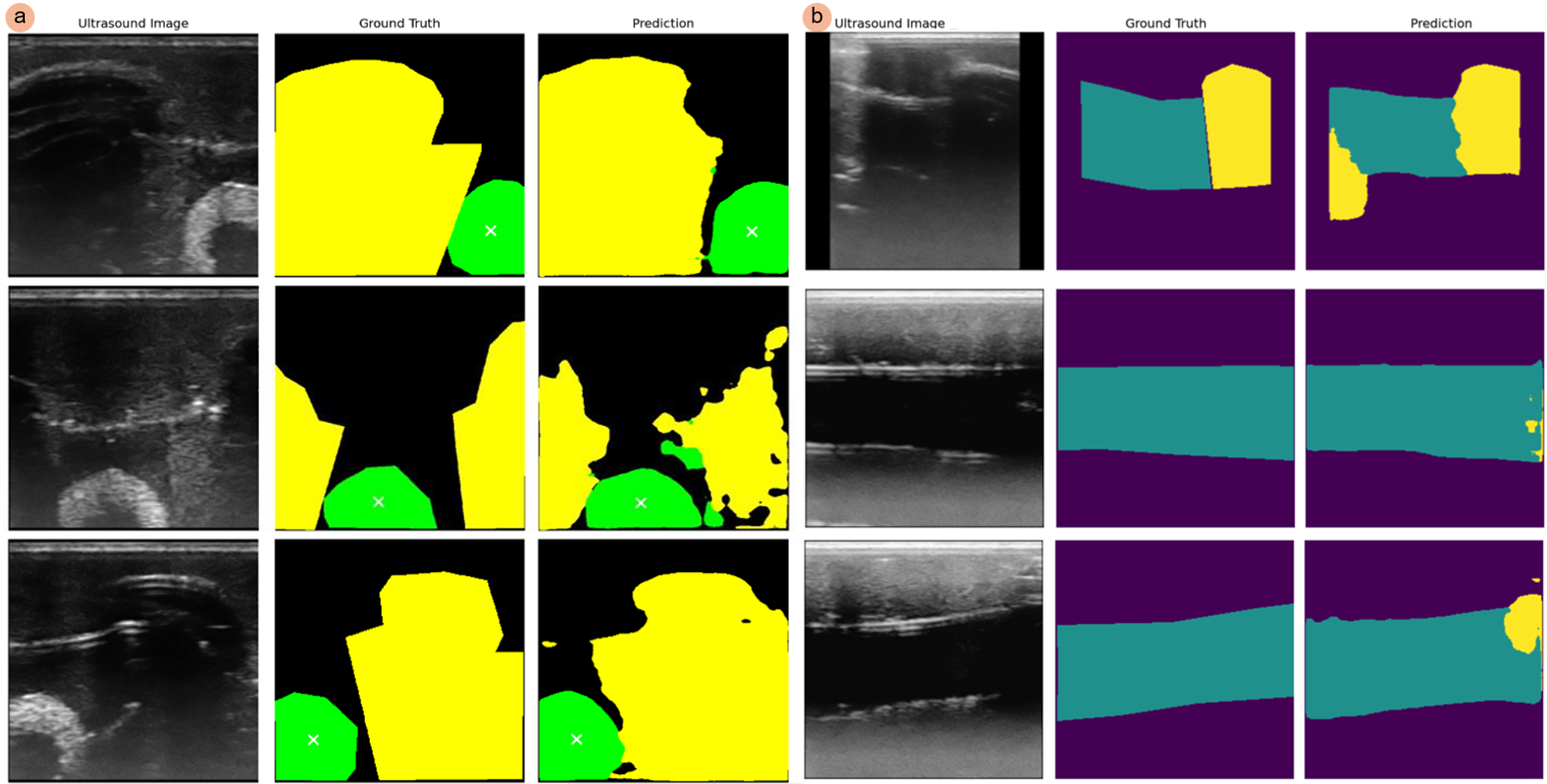}
\caption {Representative US segmentation results:
(a) tissue and bone segmentation using U-Net++, and
(b) pleural fluid region and bone segmentation using Attention U-Net.
Columns show the B-mode image, ground truth, and predicted mask.
Bone, tissue, and pleural fluid region are shown in yellow, green, and
teal, respectively; the white cross marks the selected intervention
location.}
\label{fig:segmentation_tissue_masks}
\end{figure}

Representative results are shown in
Fig.~\ref{fig:segmentation_tissue_masks}. The predicted segmentation
masks closely correspond to the ground-truth labels for both the
procedure-specific targets and the surrounding bone structures,
supporting the mean segmentation performance reported in
Table~\ref{tab:segmentation}.

% \begin{figure}[!h]
% \centering
% \includegraphics[width=\columnwidth]{images/Pleural Effusion Segmentation.png}
% \caption {\textcolor{black}{Qualitative ultrasound segmentation results for the pleural zone and bone segmentation using
% Attention U-Net. Columns show the B-mode input, pixel-wise ground truth, and predicted mask, respectively. Bone and pleural are shown in yellow and teal colours.}}
% \label{fig:segmentation_pleural_masks}
% \end{figure}

% \begin{table}[b]
% \centering
% \caption{Procedure-specific ultrasound segmentation performance.}
% \label{tab:segmentation}
% \begin{tabular}{llcccc}
% \hline
% Procedure & Model & Target DSC & Bone DSC & mDSC & HD95 \\
% \hline
% \multirow{3}{*}{Aspiration}
%  & U-Net         & $0.879\pm 0.009 $& $0.753\pm 0.008 $ & 0.816 & 40.84 \\
%  & Att U-Net & $ 0.912\pm 0.006 $ & $ 0.721\pm 0.007 $ & 0.817& 35.22 \\
%  & U-Net++       & $ 0.912\pm 0.007$ & $ 0.577\pm 0.009 $ & 0.745 & 43.06 \\
% \hline
% \multirow{3}{*}{Biopsy}
%  & U-Net         & $0.845 \pm 0.015$ & $0.846\pm 0.019$ & 0.845 & 65.75 \\
%  & Att U-Net & $0.810 \pm 0.016$ & $0.791 \pm 0.0194$ & 0.801 & 72.56 \\
%  & U-Net++       & $0.883 \pm 0.012$ & $0.816 \pm 0.0185$ & 0.849 & 54.80 \\
% \hline
% \end{tabular}
% \end{table}

\begin{table}[!t]
\centering
\caption{Segmentation performance on the held-out test sets.
Class-wise DSC values are reported as mean$\pm$SD.}
\label{tab:segmentation}
\footnotesize
\setlength{\tabcolsep}{3.2pt}
\renewcommand{\arraystretch}{1.10}
\begin{tabular*}{\columnwidth}{
@{\extracolsep{\fill}}lcccc@{}}
\toprule
Model &
\shortstack{Target\\DSC} &
\shortstack{Bone\\DSC} &
\shortstack{Mean\\DSC} &
\shortstack{Target\\HD95} \\
\midrule

\multicolumn{5}{@{}l}{\textit{Pleural fluid aspiration}}\\
U-Net
& $0.879{\pm}0.009$
& $\mathbf{0.753{\pm}0.008}$
& $0.816$
& $40.84$ \\

Attention U-Net
& $0.912{\pm}0.006$
& $0.721{\pm}0.007$
& $\mathbf{0.817}$
& $\mathbf{35.22}$ \\

U-Net++
& $\mathbf{0.913{\pm}0.007}$
& $0.577{\pm}0.009$
& $0.745$
& $43.06$ \\

\addlinespace[2pt]
\multicolumn{5}{@{}l}{\textit{Tissue acquisition}}\\
U-Net
& $0.845{\pm}0.015$
& $\mathbf{0.846{\pm}0.019}$
& $0.846$
& $65.75$ \\

Attention U-Net
& $0.810{\pm}0.016$
& $0.791{\pm}0.019$
& $0.801$
& $72.56$ \\

U-Net++
& $\mathbf{0.883{\pm}0.012}$
& $0.816{\pm}0.019$
& $\mathbf{0.850}$
& $\mathbf{54.80}$ \\

\bottomrule
\end{tabular*}
\end{table}

\subsection{Pleural Fluid Aspiration Evaluation}

The pleural fluid aspiration experiments evaluated the complete
perception-guided intervention pipeline, including autonomous US
scanning, identification of a bone-free aspiration region, needle
deployment, vacuum-assisted fluid transport, and needle retraction.
A total of 25 independent end-to-end trials were performed. For each
trial, the validity of the detected intervention region, successful
target access, occurrence of physical rib interaction, extracted fluid
volume, aspiration duration, and completion of the needle-retraction
sequence were recorded.

Once the intervention-region criteria were satisfied for
$N_f=5$ consecutive US frames, robotic scanning was stopped while
the corresponding probe pose and compliant surface contact were
maintained. The coaxial needle assembly was then
deployed according to the aspiration sequence described in
Section~\ref{sec:system_integration}, positioning the exposed 12G
collection lumen within the identified pleural fluid region. The
peristaltic pump was subsequently activated to transport the fluid
through the shared vacuum-assisted extraction pathway.

A trial was considered successful when the needle accessed the
identified pleural fluid region without physical rib interaction,
fluid was transported through the needle and tubing to the collection
system, and the prescribed needle-retraction sequence was completed.
Successful end-to-end aspiration was achieved in 23 of the 25 trials,
corresponding to a procedural success rate of 92\%. In the remaining
two trials, incorrect US-based localisation of the intervention region
prevented valid target access. These failures occurred before
aspiration and were therefore attributable to the perception and
access stages rather than to the extraction mechanism. Once the
pleural fluid region was successfully accessed, vacuum-assisted fluid
transport and needle retraction were completed in all 23 cases
(23/23). Across the successful trials, the mean extracted fluid volume
was 280~mL, with a mean aspiration duration of 155~s and a corresponding
mean extraction rate of 108.4~mL/min. These results demonstrate reliable
fluid acquisition and transport following valid US-guided target
access. The complete procedural outcomes are summarised in
Table~\ref{tab:pleural_aspiration}.

\begin{table}[!t]
\centering
\caption{Pleural aspiration performance over 25 trials}
\label{tab:pleural_aspiration}
\begin{tabular}{lc}
\hline
\textbf{Metric} & \textbf{Result} \\
\hline
% Number of trials                 & 25 \\
Fluid extraction success          & 23/25 (92\%) \\
US localisation failure           & 2/25 (8\%) \\
Extraction following valid access & 23/23 (100\%) \\
Mean extracted fluid volume       & 280~mL \\
Mean aspiration duration          & 155~s \\
Mean extraction rate              & 108.4~mL/min \\
\hline
\end{tabular}
\end{table}

\subsection{Tissue Biopsy Evaluation}

The tissue acquisition experiments evaluated end-to-end execution of
the solid-material extraction configuration, comprising tissue
localization, bone-constrained target access, procedure-specific needle
actuation, tissue acquisition, vacuum-assisted sample transport, and
needle retraction. Twenty-five trials were performed using tissue
targets embedded within the phantom. For each trial, target detection,
bone interaction during deployment, successful tissue acquisition,
sample transport through the extraction pathway, and completion of the
needle retraction sequence were recorded. A trial was classified as successful when the tissue target was correctly identified, the needle assembly reached the target without
physical bone interaction, and acquired tissue was transported through
the 12G collection lumen and tubing to the collection system. Complete
end-to-end tissue acquisition was achieved in 22 of 25 trials,
corresponding to a procedural success rate of 88\%.

The three unsuccessful trials were further classified according to the
stage of failure. Two trials (8\%) involved bone interaction during
needle deployment, preventing successful target access, while one trial (4\%) was associated with incorrect tissue localization prior to needle deployment. No subsequent acquisition or transport failure was observed among the 22 trials in which valid target access was achieved.
Accordingly, tissue acquisition and transport following successful
target access were achieved in 22/22 cases. The experimental outcomes of tissue acquisition are summarized in Table~\ref{tab:tissue_biopsy}.

\begin{table}[!h]
\centering
\caption{Tissue biopsy performance over 25 trials.}
\label{tab:tissue_biopsy}
\begin{tabular}{lc}
\hline
\textbf{Metric} & \textbf{Result} \\
\hline
% Number of trials                 & 25 \\
Successful tissue extraction     & 22/25 (88\%) \\
Bone collision failures          & 2/25 (8\%) \\
Target detection failures        & 1/25 (4\%) \\
\hline
\end{tabular}
\end{table}

\subsection{Integrated Performance and Failure Analysis}

\begin{figure}[!t]
\centering
\includegraphics[width=\columnwidth]{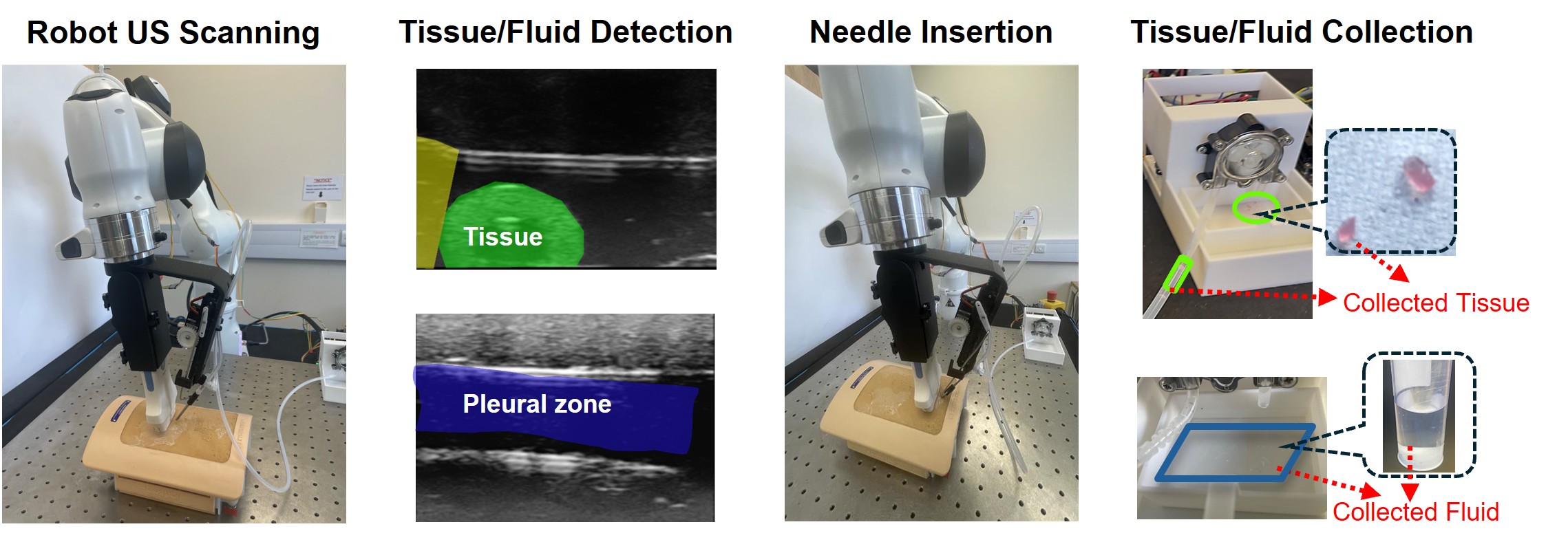}
\caption{Representative screenshots from the tissue-biopsy and pleural fluid aspiration sequence: robotic US scanning, segmentation of the tissue target and pleural effusion zone, target-guided deployment
of the coaxial needle assembly, and pump-assisted transport of the
acquired tissue and pleural fluid to the collection chamber.}
\label{fig:res2}
\end{figure}

Representative stages of the pleural fluid aspiration and tissue biopsy experiments are shown in Fig.~\ref{fig:res2}.
Across both intervention configurations, the complete end-to-end
procedure was successfully executed in 45 of the 50 trials, yielding
an overall success rate of 90\%. To determine whether the remaining
failures originated from the shared robotic pipeline or the
procedure-specific extraction mechanisms, each unsuccessful trial was
assigned to the stage at which procedural progression was interrupted.

Three of the five unsuccessful trials originated from US-based target
localisation: two during pleural fluid aspiration and one during tissue
biopsy. The remaining two failures occurred during tissue-biopsy needle
deployment because of physical interaction with bone. Thus, all five
failures occurred before the target material could be acquired. In
contrast, once valid target access was achieved, material acquisition
and vacuum-assisted transport were successful in all 45 cases:
23/23 for pleural fluid aspiration and 22/22 for tissue biopsy. No
failure was observed during cannula actuation, pump-assisted material
transport, or needle retraction.

This stage-wise analysis indicates that the shared extraction pathway
reliably handled materials with substantially different physical
properties once target access had been established. It also localises
the principal remaining limitations to US-based intervention-region
selection and collision-free needle deployment, rather than to the
procedure-specific acquisition sequences. These findings support the
proposed modular architecture, in which robotic scanning, anatomical
perception, needle deployment, and material transport are shared,
while only the distal cannula actuation is adapted to the material
being acquired.

\section{Conclusion}

This work presents a unified robotic US-guided framework for pleural
fluid aspiration and tissue biopsy using a common end effector and
material-transport pathway. The system integrates force-controlled
robotic scanning, procedure-specific US perception, bone-aware
access-region selection, motorized deployment of a coaxial needle
assembly, and pump-assisted extraction. Relative actuation of the
outer cutting and inner collection cannulas accommodates the distinct
acquisition requirements of liquid and solid material without
exchanging the end effector. 

End-to-end phantom experiments achieved successful extraction in
23/25 pleural fluid aspiration trials (92\%) and 22/25 tissue biopsy
trials (88\%). These results demonstrate the feasibility of progressing
autonomously from US-guided target identification to the acquisition
and transport of both liquid and solid material using a single robotic
platform, with the potential to reduce repeated needle insertions and
associated patient discomfort.

Future work will focus on integrating additional sensing modalities at
the needle, to detect tissue transitions and support real-time adaptation to respiratory motion and
organ displacement.
The perception pipeline will be extended with temporal consistency
and uncertainty-aware deployment criteria, while AI-based needle
guidance will be investigated to compensate for tissue deformation
and physiological motion. Further development will include
end-effector miniaturization, sterilisable patient-contact
components, and validation using dynamic phantoms and ex vivo tissue.

% \section*{Acknowledgments}
% % This should be a simple paragraph before the References to thank those individuals and institutions who have supported your work on this article.

\bibliographystyle{IEEEtran}
\bibliography{References}

\vfill

\end{document}